\documentclass[10pt,twocolumn,letterpaper]{article}

\usepackage{cvpr}              % To produce the CAMERA-READY version
\usepackage{graphicx}
\usepackage{amsmath}
\usepackage{amssymb}
\usepackage{booktabs}
\usepackage{multirow}
\usepackage{multicol}
\usepackage{listings}
\usepackage{numprint}
\usepackage{xcolor}

\definecolor{codegreen}{rgb}{0,0.6,0}
\definecolor{codegray}{rgb}{0.5,0.5,0.5}
\definecolor{codepurple}{rgb}{0.58,0,0.82}
\definecolor{backcolour}{rgb}{0.95,0.95,0.92}

\lstdefinestyle{mystyle}{
    commentstyle=\color{codegreen},
    keywordstyle=\color{magenta},
    numberstyle=\tiny\color{codegray},
    stringstyle=\color{codepurple},
    basicstyle=\ttfamily\footnotesize,
    breakatwhitespace=false,         
    breaklines=true,                 
    captionpos=b,                    
    keepspaces=true,                 
    numbers=left,                    
    numbersep=5pt,                  
    showspaces=false,                
    showstringspaces=false,
    showtabs=false,                  
    tabsize=2
}

\usepackage[pagebackref,breaklinks,colorlinks]{hyperref}

\usepackage[capitalize]{cleveref}
\crefname{section}{Sec.}{Secs.}
\Crefname{section}{Section}{Sections}
\Crefname{table}{Table}{Tables}
\crefname{table}{Tab.}{Tabs.}

\def\cvprPaperID{*****} % *** Enter the CVPR Paper ID here
\def\confName{CVPR}
\def\confYear{2022}

\begin{document}

\title{\includegraphics[width=.035\linewidth]{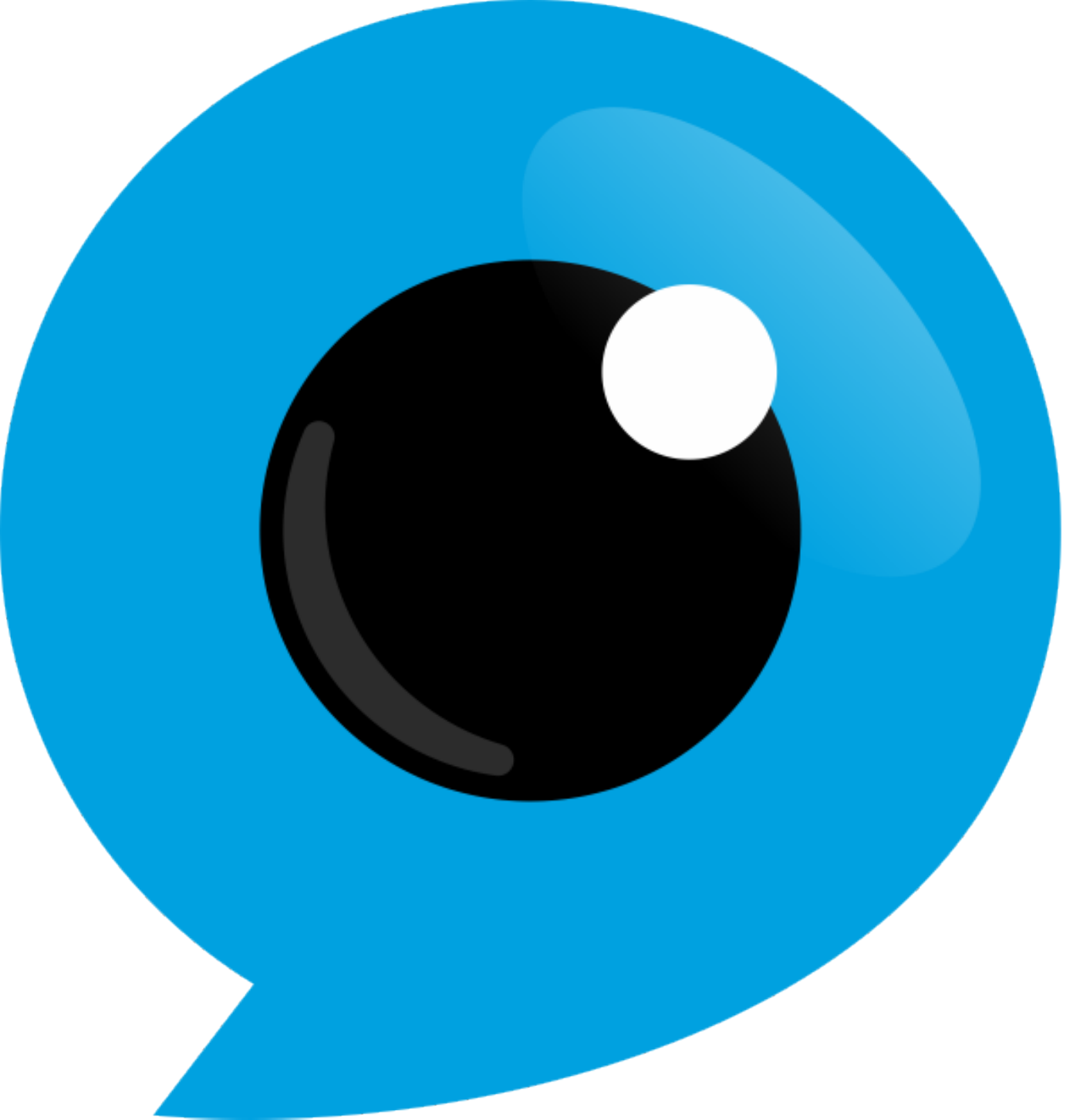}~LAVIS: A Library for Language-Vision Intelligence}

% \and
\author{Dongxu Li\textsuperscript{*}, Junnan Li\textsuperscript{*}, Hung Le, Guangsen Wang, Silvio Savarese, Steven C.H. Hoi\textsuperscript{*}\\
\\
%{\tt\small\{li.d,junnan.li,shoi\}@salesforce.com}\\
Salesforce Research}
\maketitle

\def\thefootnote{*}\footnotetext{Correspondence: \texttt{\{li.d,junnan.li,shoi\}@salesforce.com}}

%%%%%%%%% ABSTRACT
\begin{abstract}
We introduce LAVIS, an open-source deep learning library for LAnguage-VISion research and applications.
LAVIS aims to serve as a one-stop comprehensive library that brings recent advancements in the language-vision field accessible for researchers and practitioners, as well as fertilizing future research and development.
It features a unified interface to easily access state-of-the-art image-language, video-language models and common datasets. LAVIS supports training, evaluation and benchmarking on a rich variety of tasks, including multimodal classification, retrieval, captioning, visual question answering, dialogue and pre-training.
%
% In addition, LAVIS provides useful toolkit to help download and browse built-in datasets to help quickly prepare and better understand the multimodal data.
%
In the meantime, the library is also highly extensible and configurable, facilitating future development and customization.
In this technical report, we describe design principles, key components and functionalities of the library, and also present benchmarking results across common language-vision tasks.
The library is available at: \emph{\textbf{\url{https://github.com/salesforce/LAVIS}}}.

\end{abstract}

%%%%%%%%% BODY TEXT
\section{Introduction}
\label{sec:intro}
\begin{figure}[h!]    
    \begin{center}{\includegraphics[width=0.96\linewidth]{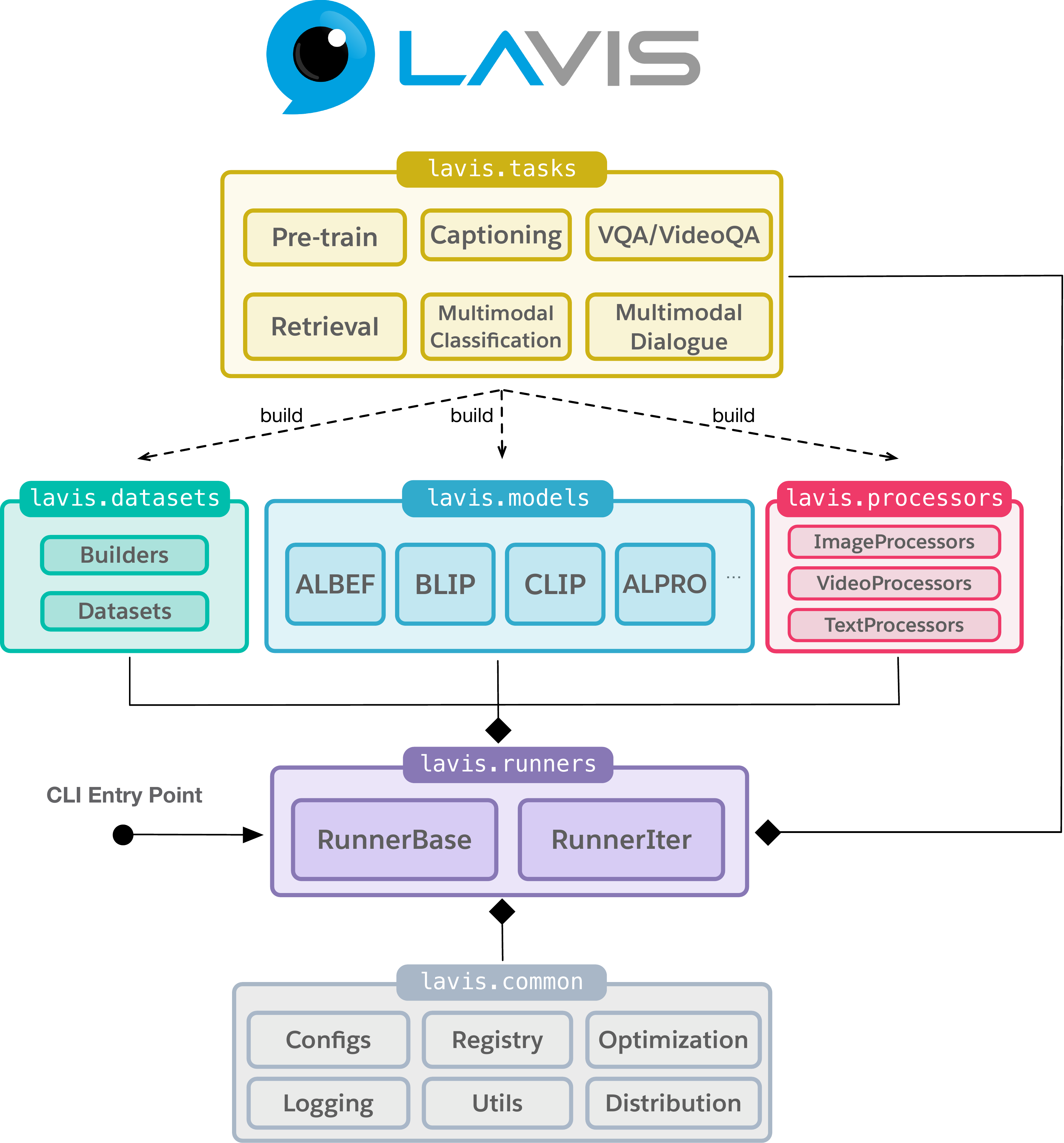}
    \caption{Overall architecture of the LAVIS library.}\label{fig:arch}}\end{center}
    \vspace{-1.3em}
\end{figure}

Multimodal content, in particular language-vision data including texts, images and videos are ubiquitous for real-world applications, such as content recommendation, e-commerce and entertainment.
There has been tremendous recent progress in developing powerful language-vision models\cite{VL-BERT,ViLBERT,uniter,oscar,soho,ALBEF,clip,VLP,villa,VL_T5,vinvl,BLIP,zhu2020actbert,fit,xu2021videoclip,lei2021less,alpro}.
However, training and evaluating these models across tasks and datasets require domain knowledge and are not always welcoming to incoming researchers and practitioners.
This is mainly due to inconsistent interfaces across models, datasets and task evaluations, and also the duplicating yet non-trivial efforts to prepare the required experiment setup.
To make accessible the emerging language-vision intelligence and capabilities to a wider audience, promote their practical adoptions, and reduce repetitive efforts in future development, we build LAVIS (short for LAnguage-VISion), an open-source library for training, evaluating state-of-the-art language-vision models on a rich family of common tasks and datasets, as well as for off-the-shelf inference on customized language-vision data.

Figure~\ref{fig:arch} shows the overall design of LAVIS. 
Important features of LAVIS include (i) \textbf{Unified interface and modular design}. Key components in the library are organized using a unified and modular design. This allows effortless off-the-shelf access to individual components, swift development and easy integration of new or external components. The modular design also eases model inferences, such as multimodal feature extraction.
(ii) \textbf{Comprehensive support of image-text, video-text tasks and datasets}. LAVIS supports a growing list of more than ten common language-vision tasks, across over 20 public datasets. These tasks and datasets provide a comprehensive and unified benchmark for evaluating language-vision models.
(iii) \textbf{State-of-the-art and reproducible language-vision models}. The library enables access to over 30 pre-trained and task-specific fine-tuned model checkpoints of four foundation models: ALBEF\cite{ALBEF}, BLIP\cite{BLIP}, CLIP\cite{clip} and ALPRO\cite{alpro}. These models achieve competitive performance across multiple tasks evaluated using common metrics. We also provide training, evaluation scripts and configurations to facilitate reproducible language-vision research and adoption.
(iv) \textbf{Resourceful and useful toolkit}. In addition to the core library functionalities, we also provide useful resources to further reduce the learning barriers for the language-vision research. This includes automatic dataset downloading tools to help prepare the supported datasets, a GUI dataset browser to help preview downloaded datasets and dataset cards documenting sources, supported tasks, common metrics and leaderboards.

\section{Related Work}
\begin{table*}[!t]
% \resizebox{0.96\textwidth}{!}
\centering
{
\caption{Comparison of features in LAVIS and other existing language-vision libraries or codebase. Note that language-vision models in UniLM and TorchMultimodal (alpha release) are under development, therefore, the table only includes their supported features by the publication time of this technical report.}
\label{tab:compare}\begin{tabular}	{c c  c c c c c}
\toprule
   & & LAVIS (Ours) & MMF & UniLM & X-modaler & TorchMultimodal
   \\
   \midrule
   \multicolumn{2}{c}{Unified Model and Dataset Interface}  & \checkmark &  &  &  & \\
    \multicolumn{2}{c}{Modular Library Design}  & \checkmark & \checkmark &  & \checkmark & \checkmark \\
   \multicolumn{2}{c}{Pre-trained Model Checkpoints}  & \checkmark &  &  & &  \\
   \multicolumn{2}{c}{Task-specific Finetuned Model Checkpoints}  & \checkmark &  &  & \checkmark &  \\
   % \multicolumn{2}{c}{Distributed Training} & \checkmark & \checkmark & \checkmark & \checkmark & \checkmark \\
   \midrule
   \multirow{2}{*}{Modalities} 
   & Image-Text  & \checkmark & \checkmark&  \checkmark & \checkmark & \checkmark \\
   & Video-Text & \checkmark &  \checkmark &  &  \checkmark & \\
    \midrule
   \multirow{6}{*}{Tasks} & End2end Pre-training & \checkmark &  & \checkmark &  & \checkmark \\ 
    & Multimodal Retrieval & \checkmark & \checkmark &  &\checkmark & 
    \\ 
    & Captioning & \checkmark & \checkmark &  & \checkmark & 
    \\  
    & Visual Question Answering & \checkmark & \checkmark &  & \checkmark &  \\ 
    & Multimodal Classification & \checkmark
 & \checkmark &  &  &  \\
    & Visual Dialogue & \checkmark &  &  &  & \\    
    & Multimodal Feature Extraction & \checkmark &  &  &  & \\
    \midrule
    \multirow{5}{*}{Toolkit} 
    & Benchmarks & \checkmark &  &  &  &  \\& Dataset Auto-downloading & \checkmark & \checkmark &  &  & \\
    & Dataset Browser & \checkmark &  &  &  & \\
    & GUI Demo & \checkmark &  &  &   & \\
    & Dataset Cards & \checkmark &  &  &  &   \\
   \bottomrule
   \end{tabular}}
\end{table*}	
Table~\ref{tab:compare} summarizes the comparisons between LAVIS' key features with those of other libraries. 
Most related libraries include MMF\cite{mmf}, UniLM\cite{unilm}, X-modaler\cite{xmodaler} and TorchMultimodal\cite{torchmultimodal}.
\begin{itemize}
\itemsep0em 
\item MMF is a comprehensive multimodal framework encapsulating many language-vision models and datasets.
It implements modular interface for training and evaluation.
However, it consists of mostly task-specific architectures. Besides showing relatively inferior performance, these models are usually not easy to transfer across tasks.
Among the included foundation models\cite{VisualBERT,uniter,vinvl,ALBEF} in MMF, few fully supports finetuning or benchmarking on the extended list of downstream tasks.
In contrast, considering that pre-trained foundation models prevail across overwhelmingly many tasks and datasets with more principal and unified architectures, our library focuses on pre-trained models and their task-specific variants instead. 
\item UniLM was initiated for developing large language models, and recently also aggregates multiple standalone repositories of multimodal models.
Yet, support for multimodal models in UniLM is limited in its current development status.
Moreover, UniLM does not provide unified or modular interfaces to allow easy access or reproduction.
\item X-modaler supports a limited number of tasks and datasets, which are not as comprehensive as LAVIS. Besides, similar to MMF, models in X-modaler are also mostly in task-specific architectures. The few supported foundation model, \eg\cite{uniter}, achieves inferior results than models in LAVIS.
\item A concurrent yet in-progress\footnote{by the publication date of this report.} library TorchMultimodal\cite{torchmultimodal} promotes modular development of language-vision models.
Our library supports a wider range of tasks and datasets than TorchMultimodal while being more comprehensive and resourceful.

\end{itemize}
Other open-source implementations of individual models exist\cite{uniter,oscar,ViLBERT,clip,villa,lei2021less}, yet do not provide centralized access.
In summary, in contrast to previous efforts, our library stands out by providing \emph{easier} access to \emph{stronger} models on comprehensively \emph{many} tasks and datasets.
With this effort, we hope to significantly reduce the cost and effort to leverage and benchmark existing multimodal models, as well as to develop new models.

\section{Supported Tasks, Datasets and Models}

\begin{table*}[!t]
\centering{
\caption{Supported tasks, datasets and models in LAVIS.}
\label{tab:supported-tasks}
\resizebox{0.96\textwidth}{!}{
\begin{tabular}	{c c  c c c c c}
\toprule
\textbf{Supported Tasks} & \textbf{Supported Models} & \textbf{Supported Datasets} \\
\midrule
\multirow{2}{*}{Image-text Pre-training} & \multirow{2}{*}{ALBEF, BLIP} & \multirow{2}{*}{\shortstack{COCO, Visual Genome, SBU Caption, \\ Conceptual Captions (3M, 12M), LAION}} \\
\\
Image-text Retrieval & ALBEF, BLIP, CLIP & COCO, Flickr30k \\
Visual Question Answering & ALBEF, BLIP & VQAv2, OKVQA, A-OKVQA \\
Image Captioning & BLIP & COCO Caption, NoCaps \\
Image Classification & CLIP & ImageNet \\
Natural Language Visual Reasoning (NLVR$^2$) & ALBEF, BLIP & NLVR$^2$\\
Visual Entailment & ALBEF & SNLI-VE \\
Visual Dialogue & BLIP & VisDial \\
Video-text Retrieval & ALPRO, BLIP & MSRVTT, DiDeMo\\
Video Question Answering & ALPRO, BLIP & MSRVTT-QA, MSVD-QA\\
Video Dialogue & BLIP & AVSD \\
\bottomrule
\end{tabular}
}}
\end{table*}\label{tab:tasks}
Table~\ref{tab:tasks} summarizes the supported tasks, datasets and models in LAVIS.
In particular, we prioritize tasks that are standard, widely adopted for evaluation, and with publicly available datasets.
For image-text tasks, the library implements image-text retrieval, image captioning, visual question answering (VQA), visual dialogue, visual entailment (VE), natural language visual reasoning (NLVR$^2$) and image classification.
For video-text tasks, LAVIS currently support video-text retrieval and video question answering (VideoQA).
There are in total over 20 public datasets supported, including MSCOCO \cite{coco}, Flickr30k\cite{flickr}, VQAv2\cite{VQA2}, OK-VQA\cite{okvqa}, A-OK-VQA\cite{aokvqa}, Visual Genome\cite{VG}, ImageNet\cite{imagenet}, NoCaps\cite{nocaps}, Conceptual Captions\cite{cc,cc12m}, SBU-caption\cite{sbu}, LAION\cite{laion}, 
NLVR$^2$\cite{NLVR},
SNLI-VE\cite{SNLI},
VisDial\cite{VisDial},
AVSD\cite{avsd},
MSRVTT\cite{msrvtt}, MSVD\cite{msvd}, DiDeMo\cite{didemo} and their task-specific variants.

LAVIS currently supports 4 foundation models, \ie ALBEF\cite{ALBEF}, BLIP\cite{BLIP}, CLIP\cite{clip} and ALPRO\cite{alpro}. 
\begin{itemize}
    \item ALBEF is an image-text model. It employs a ViT\cite{vit} as the image encoder, early BERT\cite{bert} layers as the text encoder, and re-purposes late BERT layers as the multimodal encoder by adding cross-attentions. It proposes the novel image-text contrastive (ITC) loss to align unimodal features before fusing them using the multimodal encoder. It is also one of the first few models requiring no region information while demonstrating strong multimodal understanding capability.
    \item BLIP primarily tackles image-text tasks, while also showing strong zero-shot transfer capabilities to video-text tasks. It employs a ViT as the image encoder and a BERT as the text encoder. To facilitate multimodal understanding and generation, BLIP proposes mixture of encoder-decoder (MED), which re-purposes BERT into multimodal encoder and decoder with careful weight sharing. Moreover, BLIP proposes dataset bootstrapping to improve the quality of texts in the pre-training corpus by removing noisy ones and generating new diverse ones. In addition to the improved understanding capability compared to ALBEF, BLIP highlights its strong text generation ability, producing accurate and descriptive image captions. When adapted to video-text tasks, it operates on sampled frames while concatenating their features to represent the video.
    \item CLIP is a family of powerful image-text models. Different from ALBEF and BLIP, CLIP models adopt two unimodal encoders to obtain image and text representations. CLIP maximizes the similarity between positive image-text pairs, and was trained on 400M image-text pairs, rendering strong and robust unimodal representations. CLIP variants employ different visual backbones, including ResNet-50\cite{resnet}, ViT-{B/16}, ViT-{B/32}, ViT-{L/14}, ViT-{L/14-336}. We integrate a third-party implementation of CLIP\cite{openclip} into LAVIS while including the official pre-trained weights.
    \item ALPRO is a video-text model, tackling video-text retrieval and video question answering tasks. It uses TimeSformer\cite{timesformer} to extract video features, and BERT to extract text features. Similar to ALBEF, ALPRO uses contrastive loss to align unimodal features, yet it opts to use self-attention to model multimodal interaction. This architecture choice enables an additional visual-grounded pre-training task, \ie prompt entity modeling (PEM) to align fine-grained video-text information. ALPRO is strong in extracting regional video features and remains competitive for video understanding tasks across various datasets.
\end{itemize}

\section{Library Design}
This section delineates the design of LAVIS as shown in Figure~\ref{fig:arch}. Our key design principle is to provide a simple and unified library to easily (i) train and evaluate the model; (ii) access supported models and datasets; (iii) extend with new models, tasks and datasets.

\subsection{Description on each library component}
Key components in LAVIS include:
\begin{itemize}
    \item \textbf{Runners} -- \texttt{lavis.runners} module manages the overall training and evaluation lifecycle. It is also responsible for creating required components lazily as per demand, such as optimizers, learning rate schedulers and dataloaders. Currently, \texttt{RunnerBase} implements epoch-based training and \texttt{RunnerIters} implements iteration-based training.
    \item \textbf{Tasks} -- \texttt{lavis.tasks} module implements concrete training and evaluation logic per task. This includes pre-training and finetuning tasks as listed in Table~\ref{tab:tasks}. The rationale to have an abstraction of task is to accommodate task-specific training, inference and evaluation. For example, evaluating a retrieval model is different from a classification model.
    \item \textbf{Datasets} -- \texttt{lavis.datasets} module helps create datasets. Specifically, \texttt{datasets.builders} module loads dataset configurations, downloads annotations and builds the dataset;
    \begin{itemize}
        \item \texttt{lavis.datasets.datasets} module defines the supported datasets, each is a PyTorch dataset instance.
        \item We also provide automatic dataset downloading tools in \texttt{datasets/download$\_$scripts} to help prepare common public datasets.
    \end{itemize}
    \item \textbf{Models} -- \texttt{lavis.models} module holds definitions for the supported models and shared model layers.
    \item \textbf{Processors} --\texttt{lavis.processors} module handles preprocessing of multimodal input. A processor transforms input images, videos and texts into the desired form that models can consume.
    \item \textbf{Common tools and utilities} -- \texttt{lavis.commons} module contains shared classes and methods used by multiple other modules. For example, \texttt{configs} module contains classes to store and manipulate configuration files used by LAVIS. In particular, we use a hierarchical configuration design, to allow highly customizable training and evaluation. The \texttt{registry} module serves as a centralized place to manage modules that share the same functionalities. It allows building datasets, models, tasks, and learning rate schedulers during runtime, by specifying their names in the configuration; \texttt{optims} contains definitions of learning rate schedulers; \texttt{utils} contains miscellaneous utilities, mostly IO-related helper functions;
\end{itemize}

\subsection{Example library usage}
The design of the library enables easy access to existing models and future development. In this section, we include a few examples to demonstrate some common use cases.
% \vspace{-1em}
\subsubsection*{Unified interface for loading datasets and models}
LAVIS provides unified interface \texttt{load$\_$dataset} and \texttt{load$\_$model} to access supported datasets and models.
This is helpful for off-the-shelf use of datasets and model inference etc.
In the first example, we show how to load a dataset using the library.
\begin{lstlisting}[language=Python]
from lavis.datasets.builders import load_dataset
# load a specific dataset
coco_dataset = load_dataset("coco_caption")
# dataset is organized by split names.

print(coco_dataset.keys())
# dict_keys(['train', 'val', 'test'])
# total number of samples in the training split.
print(len(coco_dataset["train"]))
# 566747
# peek a random sample
print(coco_dataset["train"][0])
# {
#   'image': <PIL.Image.Image image mode=RGB size=640x480>,
#   'text_input': 'A woman wearing a net on her head cutting a cake. ',
#   'image_id': 0
# }
\end{lstlisting}
Models and their related preprocessors can also be loaded via a unified interface, which facilitates effortless analysis and inference on custom data.
In the following, we show an example that uses a BLIP captioning model to generate image captions.
\begin{lstlisting}[language=Python]
from lavis.models import load_model_and_preprocess

# load model and preprocessors
model, vis_procs, _ = load_model_and_preprocess(
  name="blip_caption", model_type="base_coco")

# raw_image is a PIL Image instance
raw_image = coco_dataset["test"][0]["image"]
# preprocess a raw input image
image = vis_procs["eval"](raw_image).unsqueeze(0)
# generate caption
caption = model.generate({"image": image})
# ['a man riding a motorcycle down a dirt road']
\end{lstlisting}
\vspace{-1em}
\subsubsection*{Unified interface for multimodal feature extraction}
LAVIS supports a unified interface to extract multimodal features. The features are useful especially for offline applications where end-to-end finetuning is not affordable. By changing \texttt{name} and \texttt{model$\_$type}, users can choose to use different model architecture and pre-trained weights.
\begin{lstlisting}[language=Python]
# load feature extraction models and processors
model, vis_procs, txt_procs = load_model_and_preprocess(
        name="blip_feature_extractor",
        model_type="base"
)
# a random instance from coco dataset
raw_image = coco_dataset["test"][0]["image"]
text = coco_dataset["test"][0]["text_input"]
# process the input
image = vis_procs["eval"](raw_image).unsqueeze(0)
text_input = txt_procs["eval"](text)
sample = {"image": image,
          "text_input": [text_input]}

# extract multimodal features
features = model.extract_features(sample)
\end{lstlisting}

\section{Benchmarks and Library Toolkit}
In this section, we benchmark model performance across tasks and datasets in LAVIS.
Then we take our web demo interface to show a few case studies on multimodal content understanding. We also present a GUI dataset browser that helps to preview supported datasets.

\subsection{Main results} The purpose of the benchmark is two-fold.
First, since most models in LAVIS are integrated from prior works, we use the benchmark to validate that our re-implementation faithfully replicates official models.
Second, the benchmark also serves as a reference for further development and extension.
In Table~\ref{tab:albef}-\ref{tab:clip},  we organize benchmark results by models and compare our replication results with those reported officially. Experiments are conducted on NVIDIA A100 GPUs.
\begin{table}[!t]
\centering{
\caption{Comparison between official and replicated task performance using ALBEF. TR denotes text retrieval; IR denotes image retrieval. The impl. columns indicates results are from official implementation (\includegraphics[width=.03\linewidth]{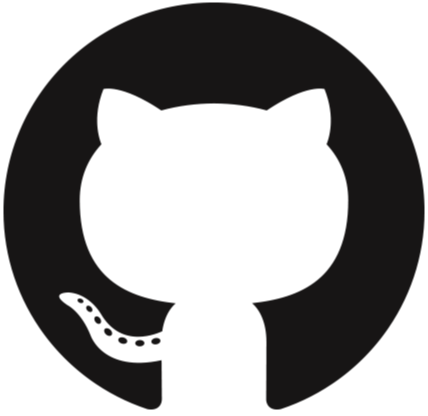}) or replication in LAVIS (\includegraphics[width=.03\linewidth]{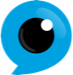}). (*) We use COCO Karpathy split\cite{karpathy} in all the experiments.}
\label{tab:albef}
\resizebox{0.48\textwidth}{!}
{
\begin{tabular}	{c c  c c c c c c}
\toprule
\textbf{Tasks} & \textbf{Datasets} & \textbf{Impl.} & \multicolumn{3}{c}{\textbf{Results}} \\
\midrule
\textbf{Retrieval}&&&\textbf{R1}&\textbf{R5}&\textbf{R10}\\
\multirow{2}{*}{TR} & \multirow{2}{*}{COCO$^{*}$} & \includegraphics[width=.03\linewidth]{images/GitHub-Mark.png} & 77.6 & 94.3 & 97.2 \\
& & \includegraphics[width=.03\linewidth]{images/logo-png.png}& 77.6 & 94.1 & 97.2 \\
\multirow{2}{*}{IR} & \multirow{2}{*}{COCO} & \includegraphics[width=.03\linewidth]{images/GitHub-Mark.png} & 60.7 & 84.3 & 90.5 \\
& & \includegraphics[width=.03\linewidth]{images/logo-png.png} & 61.0 & 84.5 & 90.7 \\
\multirow{2}{*}{TR} & \multirow{2}{*}{Flickr30k} & \includegraphics[width=.03\linewidth]{images/GitHub-Mark.png} & 95.9 & 99.8 & 100.0 \\
& & \includegraphics[width=.03\linewidth]{images/logo-png.png}& 95.8 & 99.8 & 100.0 \\
\multirow{2}{*}{IR} & \multirow{2}{*}{Flickr30k} & \includegraphics[width=.03\linewidth]{images/GitHub-Mark.png} & 85.6 & 97.5 & 98.9 \\
& & \includegraphics[width=.03\linewidth]{images/logo-png.png} & 85.5 & 97.4 & 98.9\\
\midrule
\multirow{3}{*}{\textbf{VQA}}&&& \textbf{dev} & \textbf{std}\\
& \multirow{2}{*}{VQAv2} & \includegraphics[width=.03\linewidth]{images/GitHub-Mark.png} & 75.84 & 76.04 & \\
& & \includegraphics[width=.03\linewidth]{images/logo-png.png}& 76.35 & 76.54 \\
\midrule
\multirow{6}{*}{\shortstack{\textbf{Multimodal}\\\textbf{Classification}}}
&&&\multirow{1}{*}{\textbf{val}}&\multirow{1}{*}{\textbf{test}}&\\
 & \multirow{2}{*}{SNLI-VE}  & \includegraphics[width=.03\linewidth]{images/GitHub-Mark.png} & 80.80 & 80.81\\ && 
\includegraphics[width=.03\linewidth]{images/logo-png.png} & 80.60 & 81.04\\
 & \multirow{2}{*}{NLVR2}  & \includegraphics[width=.03\linewidth]{images/GitHub-Mark.png} & 82.55 & 83.14\\ && 
\includegraphics[width=.03\linewidth]{images/logo-png.png} & 82.47 & 82.91\\
\bottomrule
\end{tabular}
}}
\end{table}
\begin{table}[!t]
\centering{
\caption{Comparison between official and replicated performance using BLIP. TR denotes text retrieval; IR denotes image retrieval. Results are produced by BLIP$_{\textnormal{CapFilt-L}}$ model. NoCaps results are reported on the entire validation set. Retrieval and captioning results are reported on the test sets; B@4 denotes BLEU-4.}
\label{tab:blip}
\resizebox{0.48\textwidth}{!}
{
\begin{tabular}	{c c  c c c c c c}
\toprule
\textbf{Tasks} & \textbf{Datasets} & \textbf{Impl.} & \multicolumn{3}{c}{\textbf{Results}} \\
\midrule
\textbf{Retrieval}&&&\textbf{R1}&\textbf{R5}&\textbf{R10}\\
\multirow{2}{*}{TR} & \multirow{2}{*}{COCO} & \includegraphics[width=.03\linewidth]{images/GitHub-Mark.png} & 82.4 & 95.4 & 97.9 \\
& & \includegraphics[width=.03\linewidth]{images/logo-png.png}& 82.0 & 95.8 & 98.1 \\
\multirow{2}{*}{IR} & \multirow{2}{*}{COCO} & \includegraphics[width=.03\linewidth]{images/GitHub-Mark.png} & 65.1 & 86.3 & 91.8 \\
& & \includegraphics[width=.03\linewidth]{images/logo-png.png} & 64.5 & 86.0 & 91.7 \\
\multirow{2}{*}{TR} & \multirow{2}{*}{Flickr30k} & \includegraphics[width=.03\linewidth]{images/GitHub-Mark.png} & 97.2 & 99.9 & 100.0 \\
& & \includegraphics[width=.03\linewidth]{images/logo-png.png}& 96.9 & 99.9 & 100.0 \\
\multirow{2}{*}{IR} & \multirow{2}{*}{Flickr30k} & \includegraphics[width=.03\linewidth]{images/GitHub-Mark.png} & 87.5 & 97.7 & 98.9 \\
& & \includegraphics[width=.03\linewidth]{images/logo-png.png} & 87.5 & 97.6 & 98.9\\
\midrule
\multirow{3}{*}{\textbf{VQA}}&&& \textbf{dev} & \textbf{std}\\
& \multirow{2}{*}{VQAv2} & \includegraphics[width=.03\linewidth]{images/GitHub-Mark.png} & 78.25 & 78.32 & \\
& & \includegraphics[width=.03\linewidth]{images/logo-png.png}& 78.23 & 78.29\\
\midrule
\multirow{5}{*}{\textbf{\shortstack{Image\\ Captioning}}}&&& \textbf{B@4} & \textbf{CIDEr} & \textbf{SPICE}\\
& \multirow{2}{*}{COCO} & \includegraphics[width=.03\linewidth]{images/GitHub-Mark.png} & 39.7 & 133.3 & -\\
& & \includegraphics[width=.03\linewidth]{images/logo-png.png}& 39.7 & 133.5 & 23.7\\
& \multirow{2}{*}{NoCaps} & \includegraphics[width=.03\linewidth]{images/GitHub-Mark.png} & - & 109.6 & 14.7 \\
& & \includegraphics[width=.03\linewidth]{images/logo-png.png} & 31.9 & 109.1 & 14.7 \\

\midrule
\multirow{3}{*}{\shortstack{\textbf{Multimodal}\\\textbf{Classification}}}
&&&\multirow{1}{*}{\textbf{val}}&\multirow{1}{*}{\textbf{test}}&\\ & \multirow{2}{*}{NLVR2}  & \includegraphics[width=.03\linewidth]{images/GitHub-Mark.png} & 82.15 & 82.24\\ && 
\includegraphics[width=.03\linewidth]{images/logo-png.png} & 82.48 & 83.25\\
% \midrule
% \multirow{3}{*}{\shortstack{\textbf{Multimodal}\\\textbf{Dialogue}}}&&&\textbf{R1}&\textbf{R5}&\textbf{MRR}\\&\multirow{2}{*}{VisDial}  & \includegraphics[width=.03\linewidth]{images/GitHub-Mark.png} & 56.44 & 85.90 & 69.41\\&&\includegraphics[width=.03\linewidth]{images/logo-png.png}\\
\bottomrule
\end{tabular}
}}
\end{table}
\begin{table}[!t]
\centering{
\caption{Comparison between official and replicated task performance using ALPRO. TR denotes video-to-text retrieval; VR denotes text-to-video retrieval.}
\resizebox{0.43\textwidth}{!}
{\label{tab:alpro}
\begin{tabular}	{c c  c c c c c c}
\toprule
\textbf{Tasks} & \textbf{Datasets} & \textbf{Impl.} & \multicolumn{3}{c}{\textbf{Results}} \\
\midrule
\textbf{Retrieval}&&&\textbf{R1}&\textbf{R5}&\textbf{R10}\\
\multirow{2}{*}{TR} & \multirow{2}{*}{MSRVTT} & \includegraphics[width=.03\linewidth]{images/GitHub-Mark.png} & 32.0 & 60.7 & 70.8  \\
& & \includegraphics[width=.03\linewidth]{images/logo-png.png}& 33.2 & 60.5 & 71.7 \\
\multirow{2}{*}{VR} & \multirow{2}{*}{MSRVTT} & \includegraphics[width=.03\linewidth]{images/GitHub-Mark.png} & 33.9 & 60.7 & 73.2 \\
& & \includegraphics[width=.03\linewidth]{images/logo-png.png} & 33.8 & 61.4 & 72.7 \\
\multirow{2}{*}{TR} & \multirow{2}{*}{DiDeMo} & \includegraphics[width=.03\linewidth]{images/GitHub-Mark.png} & 37.9 & 67.1 & 77.9 \\
& & \includegraphics[width=.03\linewidth]{images/logo-png.png}&  38.8 & 66.4 & 76.8 \\
\multirow{2}{*}{VR} & \multirow{2}{*}{DiDeMo} & \includegraphics[width=.03\linewidth]{images/GitHub-Mark.png} & 35.9 & 67.5 & 78.8 \\
& & \includegraphics[width=.03\linewidth]{images/logo-png.png} & 36.6 & 67.5 & 77.9 \\
\midrule
\multirow{5}{*}{\textbf{VideoQA}}&&& \textbf{test} \\
& \multirow{2}{*}{MSRVTT} & \includegraphics[width=.03\linewidth]{images/GitHub-Mark.png} & 42.1 & & \\& & \includegraphics[width=.03\linewidth]{images/logo-png.png}& 42.1 & \\
& \multirow{2}{*}{MSVD} & \includegraphics[width=.03\linewidth]{images/GitHub-Mark.png} & 45.9 & & \\
& & \includegraphics[width=.03\linewidth]{images/logo-png.png}& 46.0 & \\
\bottomrule
\end{tabular}
}}
\end{table}
\begin{table}[!t]
\centering{
\caption{Comparison between official and replicated performance using CLIP-ViT-L/336. Note the relative difference is possibly due to the versioning of the model weights.}
\label{tab:clip}
\resizebox{0.48\textwidth}{!}
{
\begin{tabular}	{c c  c c c c c c}
\toprule
\textbf{Tasks} & \textbf{Datasets} & \textbf{Impl.} & \multicolumn{3}{c}{\textbf{Results}} \\
\midrule
\textbf{Retrieval}&&&\textbf{R1}&\textbf{R5}&\textbf{R10}\\
\multirow{2}{*}{TR} & \multirow{2}{*}{COCO} & \includegraphics[width=.03\linewidth]{images/GitHub-Mark.png} & 58.4 & 81.5 & 88.1 \\
& & \includegraphics[width=.03\linewidth]{images/logo-png.png}& 57.2 & 80.5 & 87.8 \\
\multirow{2}{*}{IR} & \multirow{2}{*}{COCO} & \includegraphics[width=.03\linewidth]{images/GitHub-Mark.png} & 37.8 & 62.4 & 72.2 \\
& & \includegraphics[width=.03\linewidth]{images/logo-png.png} & 36.5 & 60.8 & 71.0 \\
\multirow{2}{*}{TR} & \multirow{2}{*}{Flickr30k} & \includegraphics[width=.03\linewidth]{images/GitHub-Mark.png} & 88.0 & 98.7 & 99.4 \\
& & \includegraphics[width=.03\linewidth]{images/logo-png.png}& 86.5 & 98.0 & 99.1 \\
\multirow{2}{*}{IR} & \multirow{2}{*}{Flickr30k} & \includegraphics[width=.03\linewidth]{images/GitHub-Mark.png} & 68.7 & 90.6 & 95.2 \\
& & \includegraphics[width=.03\linewidth]{images/logo-png.png} & 67.0 & 88.9 & 93.3\\
\midrule
\multirow{3}{*}{\shortstack{\textbf{Zero-shot Image}\\\textbf{Classification}}}
&&&\multirow{1}{*}{\textbf{val}}&&\\ & \multirow{2}{*}{ImageNet}  & \includegraphics[width=.03\linewidth]{images/GitHub-Mark.png} & 76.2 \\ && 
\includegraphics[width=.03\linewidth]{images/logo-png.png} & 76.5 \\

\bottomrule
\end{tabular}
}}
\end{table}

For ALBEF, BLIP and ALPRO, we re-implement their models in LAVIS based on the official repositories\footnote{Source repos: \href{https://github.com/salesforce/ALBEF}{ALBEF}, \href{https://github.com/salesforce/BLIP}{BLIP}, \href{https://github.com/salesforce/ALPRO}{ALPRO} and \href{https://github.com/mlfoundations/open_clip}{OpenClip}.} and report finetuning results using their official pre-trained weights (Table~\ref{tab:albef}-\ref{tab:alpro}).
For CLIP models, we integrate a third-party implementation\cite{openclip} and report CLIP-ViT-L/336 zero-shot inference results using the official weights\cite{clip} (Table~\ref{tab:clip}). As can be seen in the tables, our library consistently produce similar results as reported officially. 

\subsection{Additional task results with LAVIS}
\begin{table}[!t]
\centering{
\caption{Experiment results on KVQA compared with best existing methods. Due to the submission number limits, only BLIP AOKVQA result on the test split is reported.}
\label{tab:additional}
\resizebox{0.48\textwidth}{!}
{
\begin{tabular}	{c c  c c c c c c}
\toprule
\textbf{Tasks} & \textbf{Datasets} & \textbf{Models} & \multicolumn{3}{c}{\textbf{Results}} \\
\midrule
 \multirow{9}{*}{\textbf{\shortstack{KVQA}}}&&&\textbf{test}&\\& \multirow{3}{*}{OKVQA} & KAT (Single)\cite{kat} & 53.1 &\\& & KAT (Ensemble)\cite{kat} & 54.4 & \\ & & ALBEF & 54.7 & \\
&& BLIP & 55.4 & \\
&&&\textbf{val}& \textbf{test}\\ & \multirow{3}{*}{AOKVQA} &  GPV-2\cite{gpv2} & 48.6 & 40.7 \\ & & ALBEF & 54.5 & - \\
&& BLIP (VQAv2) & 53.4 & - \\
&& BLIP & 56.2 & 50.1 \\
\midrule
\multirow{5}{*}{\shortstack{\textbf{Video}\\\textbf{Dialogue}}}
& \multirow{5}{*}{AVSD} & & \multirow{1}{*}{\textbf{B@4}} & \textbf{CIDEr} & \\ 
& & MTN \cite{le-etal-2019-multimodal} & 0.410  & 1.129 & \\ 
& & PDC \cite{le2021learning} & 0.429   & 1.194 & \\ 
& & RLM \cite{li2021avsd}  &  0.459 & 1.308 & \\ 
& & VGD-GPT &  0.465 &  1.315 & \\ 
\bottomrule
\end{tabular}
}}
\end{table}
In Table~\ref{tab:additional}, we present results by adapting models in LAVIS to new tasks and datasets, on which the models were not previously reported on.
In this way, we show that our library helps to easily adapt to new tasks and datasets, while achieving competitive performance.

\textbf{Knowledge-based VQA (KVQA)}. The task of KVQA aims to measure the commonsense knowledge learnt by language-vision models, where models are asked to answer questions involving external knowledge.
To this end, state-of-the-art models \cite{kat,gpv2} resort to external knowledge base\cite{wikidata} and/or large language models\cite{gpt}.
In our experiments, we show that language-vision pre-trained models finetuned on VQAv2\cite{VQA2} show strong transfer results to KVQA datasets. With additional finetuning on KVQA datasets, further improvements are observed on both OK-VQA and AOK-VQA datasets.
As a result, our best model BLIP surpasses previous state-of-the-art by a clear margin.

\begin{figure*}[!t]
    \begin{center}{\includegraphics[width=0.98\linewidth]{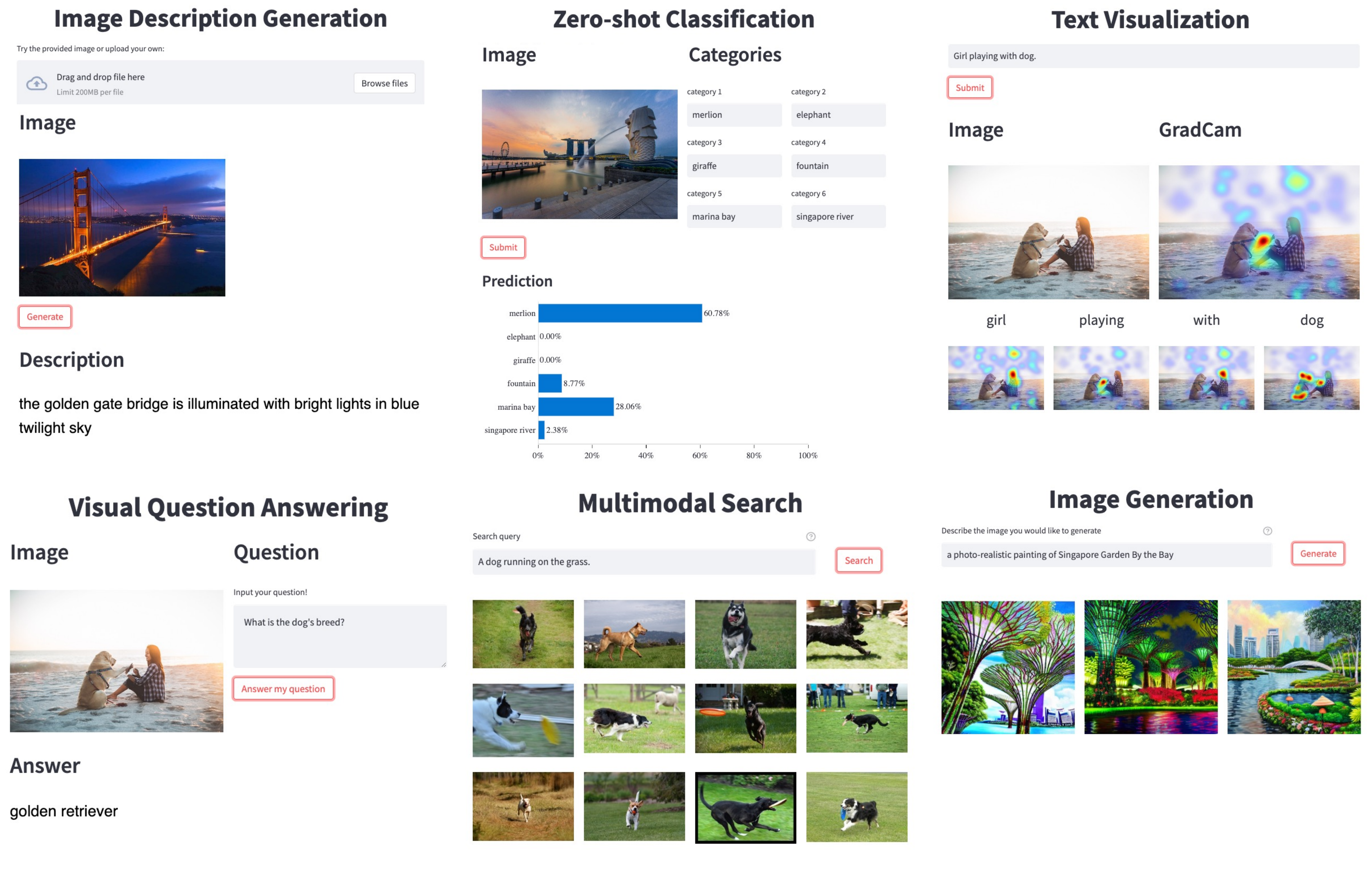}
    \caption{Screenshots of the GUI web demo.
    }\label{fig:demo}}\end{center}

\end{figure*}
% \vspace{-2em}
\begin{figure}[!t]
{\includegraphics[width=0.98\linewidth]{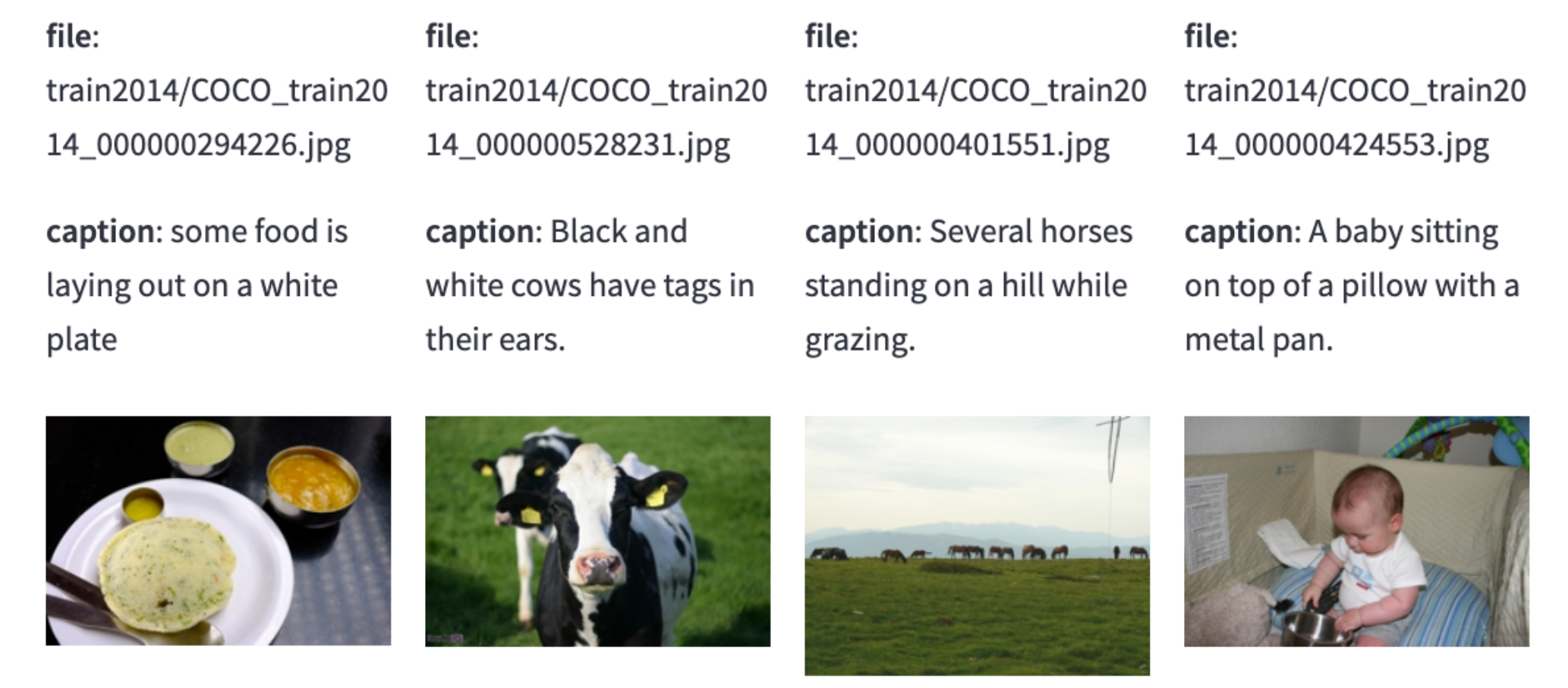}
    \caption{The developed dataset browser helps to quickly gain understanding of multimodal datasets.
    }
\label{fig:browser}}
\end{figure}

\textbf{Video Dialogue}.
The task of video-grounded dialogues requires models to generate a natural response given a dialogue context and a grounding video \cite{avsd}.
Existing models have exploited new architectural designs  \cite{le-etal-2019-multimodal}, additional learning tasks \cite{le-etal-2022-vgnmn, le2021learning}, and pretraining \cite{le-hoi-2020-video, li2021avsd} to improve the model abilities to understand multimodal context and generate natural language. 
In our experiments, we show that our library can be easily integrated with any vision-language models (such as VGD-GPT \cite{le-hoi-2020-video}) to adapt to this dialogue task. 
The results in Table \ref{tab:additional} show that our model implementation with LAVIS can lead to impressive performance, comparable to current state-of-the-art approaches. 

\subsection{Library resources and toolkit}
In addition to the components aforementioned, LAVIS also provides useful toolkit and resources to further ease development.
This includes pre-trained and finetuned model checkpoints, automatic dataset downloading tools, a web demo and a dataset browser.

\textbf{Pre-trained and finetuned model checkpoints.} We include pre-trained and finetuned model checkpoints in the library. This promotes easy replication of our experiment results and to repurpose pre-trained models for other applications. Model checkpoints are downloaded automatically upon loading models.

\textbf{Web demo}. As shown in Figure~\ref{fig:demo}, we develop a GUI-based web demo, which aims to provide a user-friendly interface to explore various multimodal capabilities. Currently the demo supports the following functionalities: (i) \emph{image captioning}: produces a caption in natural language to describe an input image; (ii) \emph{visual question answering}: answer natural language questions regarding the input image; (iii) \emph{multimodal search}: search images in a gallery given a text query; (iv) \emph{text visualization}: given an input image and a text caption, produces GradCam\cite{gradcam} for each text token on the image; (v) zero-shot multimocal classification: classify an input images into a set of input labels in text. (vi) Thanks to the modular design of LAVIS, one can easily extend the demo with new functionalities, such as \emph{text-to-image generation}, as shown in the Figure~\ref{fig:demo}. 
\textbf{Automatic dataset downloading and browsing.} Preparing language-vision datasets for pre-training and fine-tuning incurs much duplicating effort.
To this end, LAVIS provides tools to automatically download and organize the public datasets, so that users can get access to the common datasets easier and quicker.
In addition, we also develop a GUI dataset browser, as shown in Figure~\ref{fig:browser}, that helps users to rapidly gain intuitions about the data they use.

\section{Conclusion and Future Work}
We present LAVIS, an open-source deep learning library for language-vision research and applications.
The library is designed to provide researchers and practitioners with easier and comprehensive access to state-of-the-art multimodal capabilities, 
The library also features a unified interface and extensible design to promote future development.
Besides, the library also features extensive access to pre-trained weights and useful resources to reduce duplicating replication efforts.
With these features, we expect LAVIS to serve as a one-stop library in multimodal AI for a wide and growing audience.

We continue to actively develop and improve LAVIS. In future releases, our priorities are to include more language-vision models, tasks and datasets to the library. We also plan to add more parallelism support for scalable training and inference. While we will maintain LAVIS in the long term, we welcome and invite contributions from the open-source community to join this evolving effort.
\section*{Broader Impact and Responsible Use}
LAVIS can provide useful capabilities for many real-world multimodal applications.
It features easy, unified and centralized access to powerful language-vision models, facilitating effective multimodal analysis and reproducible research and development.
We encourage researchers, data scientists, and
ML practitioners to adopt LAVIS in real-world applications for positive social impacts, \eg efficient and environment-friendly large-scale multimodal analysis.

However, LAVIS may also be misused. We encourage users to read detailed discussion and guidelines for building responsible AI, \eg~\cite{responsibleai}. In particular, LAVIS should not be used to develop multimodal models that may expose unethical capabilities.

It is also important to note that that models in LAVIS provide no guarantees on their multimodal abilities; incorrect or biased predictions may be observed. In particular, the datasets and pretrained models utilized in LAVIS contain socioeconomic biases which may result in misclassification and other unwanted behaviors such as offensive or inappropriate speech.
We strongly recommend that users review the pre-trained models and overall system in LAVIS before practical adoption.
We plan to improve the library by investigating and mitigating these potential biases and inappropriate behaviors in the future.

\section*{Acknowledgement}
We thank our colleagues and leadership teams from Salesforce who have provided strong support, suggestions and contributions to this project. We also thank our ethical
AI team for their feedback.
%%%%%%%%% REFERENCES
{\small
\bibliographystyle{ieee_fullname}
\bibliography{main}

\begin{thebibliography}{10}\itemsep=-1pt

\bibitem{unilm}
Large-scale self-supervised pre-training across tasks, languages, and
  modalities.
\newblock \url{https://github.com/microsoft/unilm}, 2020.

\bibitem{torchmultimodal}
Torchmultimodal (alpha release).
\newblock \url{https://github.com/facebookresearch/multimodal}, 2022.

\bibitem{nocaps}
Harsh Agrawal, Peter Anderson, Karan Desai, Yufei Wang, Xinlei Chen, Rishabh
  Jain, Mark Johnson, Dhruv Batra, Devi Parikh, and Stefan Lee.
\newblock nocaps: novel object captioning at scale.
\newblock In {\em {ICCV}}, pages 8947--8956, 2019.

\bibitem{avsd}
Huda Alamri, Vincent Cartillier, Abhishek Das, Jue Wang, Anoop Cherian, Irfan
  Essa, Dhruv Batra, Tim~K Marks, Chiori Hori, Peter Anderson, et~al.
\newblock Audio visual scene-aware dialog.
\newblock In {\em Proceedings of the IEEE/CVF Conference on Computer Vision and
  Pattern Recognition}, pages 7558--7567, 2019.

\bibitem{didemo}
Lisa Anne~Hendricks, Oliver Wang, Eli Shechtman, Josef Sivic, Trevor Darrell,
  and Bryan Russell.
\newblock Localizing moments in video with natural language.
\newblock In {\em Proceedings of the IEEE international conference on computer
  vision}, pages 5803--5812, 2017.

\bibitem{fit}
Max Bain, Arsha Nagrani, G{\"u}l Varol, and Andrew Zisserman.
\newblock Frozen in time: A joint video and image encoder for end-to-end
  retrieval.
\newblock In {\em ICCV}, 2021.

\bibitem{responsibleai}
Kathy Baxter.
\newblock Ai is everywhere — but are you building it responsibly?
\newblock
  \url{https://www.salesforce.com/blog/build-ethical-ai/?hasLoggedIn=true},
  2022.

\bibitem{timesformer}
Gedas Bertasius, Heng Wang, and Lorenzo Torresani.
\newblock Is space-time attention all you need for video understanding?
\newblock In {\em ICML}, 2021.

\bibitem{SNLI}
Samuel~R. Bowman, Gabor Angeli, Christopher Potts, and Christopher~D. Manning.
\newblock A large annotated corpus for learning natural language inference.
\newblock In Llu{\'{\i}}s M{\`{a}}rquez, Chris Callison{-}Burch, Jian Su,
  Daniele Pighin, and Yuval Marton, editors, {\em {EMNLP}}, pages 632--642,
  2015.

\bibitem{gpt}
Tom Brown, Benjamin Mann, Nick Ryder, Melanie Subbiah, Jared~D Kaplan, Prafulla
  Dhariwal, Arvind Neelakantan, Pranav Shyam, Girish Sastry, Amanda Askell,
  et~al.
\newblock Language models are few-shot learners.
\newblock {\em Advances in neural information processing systems},
  33:1877--1901, 2020.

\bibitem{cc12m}
Soravit Changpinyo, Piyush Sharma, Nan Ding, and Radu Soricut.
\newblock {Conceptual 12M}: Pushing web-scale image-text pre-training to
  recognize long-tail visual concepts.
\newblock In {\em CVPR}, 2021.

\bibitem{uniter}
Yen{-}Chun Chen, Linjie Li, Licheng Yu, Ahmed~El Kholy, Faisal Ahmed, Zhe Gan,
  Yu Cheng, and Jingjing Liu.
\newblock {UNITER:} universal image-text representation learning.
\newblock In {\em ECCV}, volume 12375, pages 104--120, 2020.

\bibitem{VL_T5}
Jaemin Cho, Jie Lei, Hao Tan, and Mohit Bansal.
\newblock Unifying vision-and-language tasks via text generation.
\newblock {\em arXiv preprint arXiv:2102.02779}, 2021.

\bibitem{VisDial}
Abhishek Das, Satwik Kottur, Khushi Gupta, Avi Singh, Deshraj Yadav, Jos{\'{e}}
  M.~F. Moura, Devi Parikh, and Dhruv Batra.
\newblock Visual dialog.
\newblock In {\em {CVPR}}, pages 1080--1089, 2017.

\bibitem{imagenet}
Jia Deng, Wei Dong, Richard Socher, Li-Jia Li, Kai Li, and Li Fei-Fei.
\newblock Imagenet: A large-scale hierarchical image database.
\newblock In {\em 2009 IEEE conference on computer vision and pattern
  recognition}, pages 248--255. Ieee, 2009.

\bibitem{bert}
Jacob Devlin, Ming{-}Wei Chang, Kenton Lee, and Kristina Toutanova.
\newblock {BERT:} pre-training of deep bidirectional transformers for language
  understanding.
\newblock In Jill Burstein, Christy Doran, and Thamar Solorio, editors, {\em
  {NAACL}}, pages 4171--4186, 2019.

\bibitem{vit}
Alexey Dosovitskiy, Lucas Beyer, Alexander Kolesnikov, Dirk Weissenborn,
  Xiaohua Zhai, Thomas Unterthiner, Mostafa Dehghani, Matthias Minderer, Georg
  Heigold, Sylvain Gelly, Jakob Uszkoreit, and Neil Houlsby.
\newblock An image is worth 16x16 words: Transformers for image recognition at
  scale.
\newblock In {\em ICLR}, 2021.

\bibitem{villa}
Zhe Gan, Yen{-}Chun Chen, Linjie Li, Chen Zhu, Yu Cheng, and Jingjing Liu.
\newblock Large-scale adversarial training for vision-and-language
  representation learning.
\newblock In Hugo Larochelle, Marc'Aurelio Ranzato, Raia Hadsell,
  Maria{-}Florina Balcan, and Hsuan{-}Tien Lin, editors, {\em {NeurIPS}}, 2020.

\bibitem{VQA2}
Yash Goyal, Tejas Khot, Douglas Summers{-}Stay, Dhruv Batra, and Devi Parikh.
\newblock Making the {V} in {VQA} matter: Elevating the role of image
  understanding in visual question answering.
\newblock In {\em {CVPR}}, pages 6325--6334, 2017.

\bibitem{kat}
Liangke Gui, Borui Wang, Qiuyuan Huang, Alex Hauptmann, Yonatan Bisk, and
  Jianfeng Gao.
\newblock Kat: A knowledge augmented transformer for vision-and-language.
\newblock {\em arXiv preprint arXiv:2112.08614}, 2021.

\bibitem{resnet}
Kaiming He, Xiangyu Zhang, Shaoqing Ren, and Jian Sun.
\newblock Deep residual learning for image recognition.
\newblock In {\em Proceedings of the IEEE conference on computer vision and
  pattern recognition}, pages 770--778, 2016.

\bibitem{soho}
Zhicheng Huang, Zhaoyang Zeng, Yupan Huang, Bei Liu, Dongmei Fu, and Jianlong
  Fu.
\newblock Seeing out of the box: End-to-end pre-training for vision-language
  representation learning.
\newblock {\em arXiv preprint arXiv:2104.03135}, 2021.

\bibitem{openclip}
Gabriel Ilharco, Mitchell Wortsman, Ross Wightman, Cade Gordon, Nicholas
  Carlini, Rohan Taori, Achal Dave, Vaishaal Shankar, Hongseok Namkoong, John
  Miller, Hannaneh Hajishirzi, Ali Farhadi, and Ludwig Schmidt.
\newblock Openclip, July 2021.
\newblock If you use this software, please cite it as below.

\bibitem{gpv2}
Amita Kamath, Christopher Clark, Tanmay Gupta, Eric Kolve, Derek Hoiem, and
  Aniruddha Kembhavi.
\newblock Webly supervised concept expansion for general purpose vision models.
\newblock {\em arXiv preprint arXiv:2202.02317}, 2022.

\bibitem{karpathy}
Andrej Karpathy and Fei{-}Fei Li.
\newblock Deep visual-semantic alignments for generating image descriptions.
\newblock In {\em {CVPR}}, pages 3128--3137, 2015.

\bibitem{VG}
Ranjay Krishna, Yuke Zhu, Oliver Groth, Justin Johnson, Kenji Hata, Joshua
  Kravitz, Stephanie Chen, Yannis Kalantidis, Li{-}Jia Li, David~A. Shamma,
  Michael~S. Bernstein, and Li Fei{-}Fei.
\newblock Visual genome: Connecting language and vision using crowdsourced
  dense image annotations.
\newblock {\em {IJCV}}, 123(1):32--73, 2017.

\bibitem{le-etal-2022-vgnmn}
Hung Le, Nancy Chen, and Steven Hoi.
\newblock {VGNMN}: Video-grounded neural module networks for video-grounded
  dialogue systems.
\newblock In {\em Proceedings of the 2022 Conference of the North American
  Chapter of the Association for Computational Linguistics: Human Language
  Technologies}, pages 3377--3393, Seattle, United States, July 2022.
  Association for Computational Linguistics.

\bibitem{le2021learning}
Hung Le, Nancy~F. Chen, and Steven Hoi.
\newblock Learning reasoning paths over semantic graphs for video-grounded
  dialogues.
\newblock In {\em International Conference on Learning Representations}, 2021.

\bibitem{le-hoi-2020-video}
Hung Le and Steven~C.H. Hoi.
\newblock Video-grounded dialogues with pretrained generation language models.
\newblock In {\em Proceedings of the 58th Annual Meeting of the Association for
  Computational Linguistics}, pages 5842--5848, Online, July 2020. Association
  for Computational Linguistics.

\bibitem{le-etal-2019-multimodal}
Hung Le, Doyen Sahoo, Nancy Chen, and Steven Hoi.
\newblock Multimodal transformer networks for end-to-end video-grounded
  dialogue systems.
\newblock In {\em Proceedings of the 57th Annual Meeting of the Association for
  Computational Linguistics}, pages 5612--5623, Florence, Italy, July 2019.
  Association for Computational Linguistics.

\bibitem{lei2021less}
Jie Lei, Linjie Li, Luowei Zhou, Zhe Gan, Tamara~L Berg, Mohit Bansal, and
  Jingjing Liu.
\newblock Less is more: Clipbert for video-and-language learning via sparse
  sampling.
\newblock In {\em {CVPR}}, pages 7331--7341, 2021.

\bibitem{alpro}
Dongxu Li, Junnan Li, Hongdong Li, Juan~Carlos Niebles, and Steven~C.H. Hoi.
\newblock Align and prompt: Video-and-language pre-training with entity
  prompts.
\newblock In {\em CVPR}, 2022.

\bibitem{BLIP}
Junnan Li, Dongxu Li, Caiming Xiong, and Steven Hoi.
\newblock Blip: Bootstrapping language-image pre-training for unified
  vision-language understanding and generation.
\newblock {\em arXiv preprint arXiv:2201.12086}, 2022.

\bibitem{ALBEF}
Junnan Li, Ramprasaath~R. Selvaraju, Akhilesh~Deepak Gotmare, Shafiq Joty,
  Caiming Xiong, and Steven Hoi.
\newblock Align before fuse: Vision and language representation learning with
  momentum distillation.
\newblock In {\em NeurIPS}, 2021.

\bibitem{VisualBERT}
Liunian~Harold Li, Mark Yatskar, Da Yin, Cho{-}Jui Hsieh, and Kai{-}Wei Chang.
\newblock Visualbert: {A} simple and performant baseline for vision and
  language.
\newblock {\em arXiv preprint arXiv:1908.03557}, abs/1908.03557, 2019.

\bibitem{oscar}
Xiujun Li, Xi Yin, Chunyuan Li, Pengchuan Zhang, Xiaowei Hu, Lei Zhang, Lijuan
  Wang, Houdong Hu, Li Dong, Furu Wei, Yejin Choi, and Jianfeng Gao.
\newblock Oscar: Object-semantics aligned pre-training for vision-language
  tasks.
\newblock In {\em {ECCV}}, pages 121--137, 2020.

\bibitem{xmodaler}
Yehao Li, Yingwei Pan, Jingwen Chen, Ting Yao, and Tao Mei.
\newblock X-modaler: A versatile and high-performance codebase for cross-modal
  analytics.
\newblock In {\em Proceedings of the 29th ACM International Conference on
  Multimedia}, pages 3799--3802, 2021.

\bibitem{li2021avsd}
Zekang Li, Zongjia Li, Jinchao Zhang, Yang Feng, and Jie Zhou.
\newblock Bridging text and video: A universal multimodal transformer for
  audio-visual scene-aware dialog.
\newblock {\em IEEE/ACM Trans. Audio, Speech and Lang. Proc.}, 29:2476–2483,
  jan 2021.

\bibitem{coco}
Tsung{-}Yi Lin, Michael Maire, Serge~J. Belongie, James Hays, Pietro Perona,
  Deva Ramanan, Piotr Doll{\'{a}}r, and C.~Lawrence Zitnick.
\newblock Microsoft {COCO:} common objects in context.
\newblock In David~J. Fleet, Tom{\'{a}}s Pajdla, Bernt Schiele, and Tinne
  Tuytelaars, editors, {\em {ECCV}}, volume 8693, pages 740--755, 2014.

\bibitem{ViLBERT}
Jiasen Lu, Dhruv Batra, Devi Parikh, and Stefan Lee.
\newblock Vilbert: Pretraining task-agnostic visiolinguistic representations
  for vision-and-language tasks.
\newblock In Hanna~M. Wallach, Hugo Larochelle, Alina Beygelzimer, Florence
  d'Alch{\'{e}}{-}Buc, Emily~B. Fox, and Roman Garnett, editors, {\em NeurIPS},
  pages 13--23, 2019.

\bibitem{okvqa}
Kenneth Marino, Mohammad Rastegari, Ali Farhadi, and Roozbeh Mottaghi.
\newblock Ok-vqa: A visual question answering benchmark requiring external
  knowledge.
\newblock In {\em Proceedings of the IEEE/cvf conference on computer vision and
  pattern recognition}, pages 3195--3204, 2019.

\bibitem{sbu}
Vicente Ordonez, Girish Kulkarni, and Tamara~L. Berg.
\newblock Im2text: Describing images using 1 million captioned photographs.
\newblock In John Shawe{-}Taylor, Richard~S. Zemel, Peter~L. Bartlett, Fernando
  C.~N. Pereira, and Kilian~Q. Weinberger, editors, {\em {NIPS}}, pages
  1143--1151, 2011.

\bibitem{flickr}
Bryan~A. Plummer, Liwei Wang, Chris~M. Cervantes, Juan~C. Caicedo, Julia
  Hockenmaier, and Svetlana Lazebnik.
\newblock Flickr30k entities: Collecting region-to-phrase correspondences for
  richer image-to-sentence models.
\newblock In {\em {ICCV}}, pages 2641--2649, 2015.

\bibitem{clip}
Alec Radford, Jong~Wook Kim, Chris Hallacy, Aditya Ramesh, Gabriel Goh,
  Sandhini Agarwal, Girish Sastry, Amanda Askell, Pamela Mishkin, Jack Clark,
  et~al.
\newblock Learning transferable visual models from natural language
  supervision.
\newblock {\em arXiv preprint arXiv:2103.00020}, 2021.

\bibitem{laion}
Christoph Schuhmann, Richard Vencu, Romain Beaumont, Robert Kaczmarczyk,
  Clayton Mullis, Aarush Katta, Theo Coombes, Jenia Jitsev, and Aran
  Komatsuzaki.
\newblock Laion-400m: Open dataset of clip-filtered 400 million image-text
  pairs.
\newblock {\em arXiv preprint arXiv:2111.02114}, 2021.

\bibitem{gradcam}
Ramprasaath~R. Selvaraju, Michael Cogswell, Abhishek Das, Ramakrishna Vedantam,
  Devi Parikh, and Dhruv Batra.
\newblock Grad-cam: Visual explanations from deep networks via gradient-based
  localization.
\newblock In {\em {ICCV}}, pages 618--626, 2017.

\bibitem{cc}
Piyush Sharma, Nan Ding, Sebastian Goodman, and Radu Soricut.
\newblock Conceptual captions: {A} cleaned, hypernymed, image alt-text dataset
  for automatic image captioning.
\newblock In Iryna Gurevych and Yusuke Miyao, editors, {\em {ACL}}, pages
  2556--2565, 2018.

\bibitem{aokvqa}
Violetta Shevchenko, Damien Teney, Anthony Dick, and Anton van~den Hengel.
\newblock Reasoning over vision and language: Exploring the benefits of
  supplemental knowledge.
\newblock {\em arXiv preprint arXiv:2101.06013}, 2021.

\bibitem{mmf}
Amanpreet Singh, Vedanuj Goswami, Vivek Natarajan, Yu Jiang, Xinlei Chen, Meet
  Shah, Marcus Rohrbach, Dhruv Batra, and Devi Parikh.
\newblock Mmf: A multimodal framework for vision and language research.
\newblock \url{https://github.com/facebookresearch/mmf}, 2020.

\bibitem{VL-BERT}
Weijie Su, Xizhou Zhu, Yue Cao, Bin Li, Lewei Lu, Furu Wei, and Jifeng Dai.
\newblock Vl-bert: Pre-training of generic visual-linguistic representations.
\newblock In {\em {ICLR}}, 2020.

\bibitem{NLVR}
Alane Suhr, Stephanie Zhou, Ally Zhang, Iris Zhang, Huajun Bai, and Yoav Artzi.
\newblock A corpus for reasoning about natural language grounded in
  photographs.
\newblock In Anna Korhonen, David~R. Traum, and Llu{\'{\i}}s M{\`{a}}rquez,
  editors, {\em {ACL}}, pages 6418--6428, 2019.

\bibitem{wikidata}
Denny Vrande{\v{c}}i{\'c} and Markus Kr{\"o}tzsch.
\newblock Wikidata: a free collaborative knowledgebase.
\newblock {\em Communications of the ACM}, 57(10):78--85, 2014.

\bibitem{msvd}
Dejing Xu, Zhou Zhao, Jun Xiao, Fei Wu, Hanwang Zhang, Xiangnan He, and Yueting
  Zhuang.
\newblock Video question answering via gradually refined attention over
  appearance and motion.
\newblock In {\em Proceedings of the ACM international conference on
  Multimedia}, pages 1645--1653, 2017.

\bibitem{xu2021videoclip}
Hu Xu, Gargi Ghosh, Po-Yao Huang, Dmytro Okhonko, Armen Aghajanyan, Florian
  Metze, Luke Zettlemoyer, and Christoph Feichtenhofer.
\newblock Videoclip: Contrastive pre-training for zero-shot video-text
  understanding.
\newblock In {\em EMNLP}, pages 6787--6800, 2021.

\bibitem{msrvtt}
Jun Xu, Tao Mei, Ting Yao, and Yong Rui.
\newblock Msr-vtt: A large video description dataset for bridging video and
  language.
\newblock In {\em Proceedings of the IEEE conference on computer vision and
  pattern recognition}, pages 5288--5296, 2016.

\bibitem{vinvl}
Pengchuan Zhang, Xiujun Li, Xiaowei Hu, Jianwei Yang, Lei Zhang, Lijuan Wang,
  Yejin Choi, and Jianfeng Gao.
\newblock Vinvl: Making visual representations matter in vision-language
  models.
\newblock {\em arXiv preprint arXiv:2101.00529}, 2021.

\bibitem{VLP}
Luowei Zhou, Hamid Palangi, Lei Zhang, Houdong Hu, Jason~J. Corso, and Jianfeng
  Gao.
\newblock Unified vision-language pre-training for image captioning and {VQA}.
\newblock In {\em {AAAI}}, pages 13041--13049, 2020.

\bibitem{zhu2020actbert}
Linchao Zhu and Yi Yang.
\newblock Actbert: Learning global-local video-text representations.
\newblock In {\em CVPR}, pages 8746--8755, 2020.

\end{thebibliography}
}

\end{document}